# Sign Language Tutoring Tool

Oya Aran[1], Ismail Ari[1], Alexandre Benoit[2], Ana Huerta Carrillo[3], François-Xavier Fanard[4],
Pavel Campr[5], Lale Akarun[1], Alice Caplier[2], Michele Rombaut[2] and Bulent Sankur[1]

[1]Bogazici University, [2]LIS_INPG, [3]Technical University of Madrid, [4]Universite Catholique de Louvain, [5] University of West Bohemia in Pilsen

*Abstract*—In this project, we have developed a sign language tutor that lets users learn isolated signs by watching recorded videos and by trying the same signs. The system records the user's video and analyses it. If the sign is recognized, both verbal and animated feedback is given to the user. The system is able to recognize complex signs that involve both hand gestures and head movements and expressions. Our performance tests yield a 99% recognition rate on signs involving only manual gestures and 85% recognition rate on signs that involve both manual and non manual components, such as head movement and facial expressions.

## I. INTRODUCTION

THE purpose of this project is to develop a Sign Language Tutoring Demonstrator that lets users practice demonstrated signs and get feedback about their performance. In a learning step, a video of a specific sign is demonstrated to the user and in the practice step, the user is asked to repeat the sign. An evaluation of produced gesture is given to the learner; together with a synthesized version of the sign that lets the user get visual feedback in a caricatured form.

The specificity of Sign Language is that the whole message is contained not only in hand gestures and shapes (manual signs) but also in facial expressions and head/shoulder motion (non-manual signs). As a consequence, the language is intrinsically multimodal. In order to solve the hand trajectory recognition problem, Hidden Markov Models have been used extensively for the last decade. Lee and Kim [1] propose a method for online gesture spotting using HMMs. Starner et al. [2] used HMMs for continuous American Sign Language recognition. The vocabulary contains 40 signs and the sentence structure to be recognized was constrained to personal pronoun, verb, noun, and adjective. In 1997, Vogler and Metaxas [3] proposed a system for both isolated and continuous ASL recognition sentences with a 53-sign vocabulary. In a later study [4] the same authors attacked the scalability problem and proposed a method for the parallel modeling of the phonemes within an HMM framework. Most systems of Sign Language recognition concentrate on hand gesture analysis only. In , a survey on automatic sign language analysis is given and integrating non-manual signs with hand gestures is examined.

A preliminary version of the tutor we propose to develop, demonstrated at EUSIPCO, uses only hand trajectory based gesture recognition [6]. The signs selected were signs that could be recognized based on solely the trajectory of one hand. In this project, we aim at developing a tutoring system able to cope with two sources of information: hand gestures and head motion. The database contains complex signs that are performed with two hands and head gestures. Therefore, our Sign Language Recognition system fuses the data coming from two sources of information to recognize a performed sign: The shape and trajectory of the two hands and the head movements.

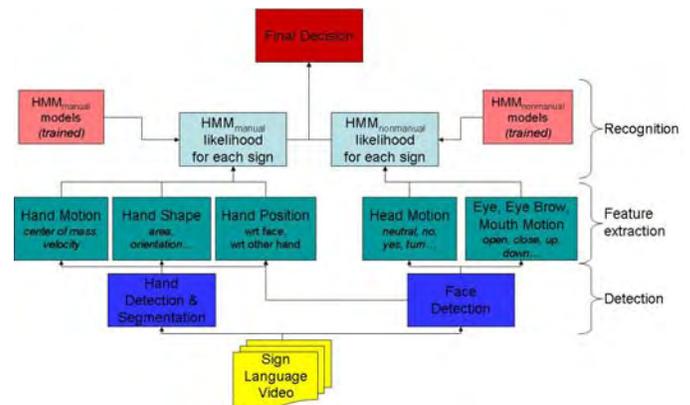

Fig. 1. Sign language recognition system block diagram

Fig. 1 illustrates the steps in sign recognition. The first step in hand gesture recognition is to detect and track both hands. This is a complex task because the hands may occlude each other and also overlap other skin colored regions, such as the arms and the face. To make the detection problem easier, markers on the hand and fingers are widely used in the literature. In this project, we have used differently colored gloves worn on the two hands. Once the hands are detected, a complete hand gesture recognition system must be able to extract the hand shape, and the hand motion. We have extracted simple hand shape features and combined them with hand motion and position information to obtain a combined feature vector. A left-to-right continuous HMM model with no state skips is trained for each sign. These HMM models could be directly used for recognition if we were to recognize only the manual signs. However, some signs involve non-manual components. Thus further analysis of head movements and facial expressions must be done to recognize non-manual signs.

Head movement analysis works concurrently with hand gesture analysis. Following the face detection step, a method based on the human visual system is used to calculate the motion energy and the velocity of the head, eye, eyebrows and mouth. These features are combined into a single feature vector and HMM models for the non-manual signs are trained.

For the final decision, manual and non-manual HMM models are fused in a sequential manner. Decisions of the manual HMMs are used as the base for decision and non-manual HMMs take part to differentiate between the variants of the base sign.

Another new feature of the Sign Language Tutoring tool is that it uses synthesized head and arm motions based on the analysis of arm and head movements. This lets the user get accentuated feedback. Feedback, either TRUE or FALSE, is given for the manual component as well as for the non-manual one, separately.

In this project, we have first defined a limited number of signs that can be used for sign language tutoring. 19 signs have been selected so that head motions are crucial for their recognition: Some signs have identical hand motions but different head motions. After defining the dataset, we have collected data from eight subjects. Each subject performed all the signs five times.

The sign language tutor application was designed to show selected signs to the user and to let the user record his/her own sign using the webcam connected to the system. The application then runs the analysis, recognition, and synthesis subsystems. The recognized sign is identified by a text message and the synthesized animation is shown as feedback to the user. If the sign is not performed correctly, the user may repeat the test.

This report is organized as follows: In section II, we give details of the sign language tutor application, together with database details. Section III details the analysis: hand segmentation, hand motion feature extraction, hand shape feature extraction, and head motion feature extraction. Section IV describes the recognition by fusion of information from all sources. Section V describes the synthesis of head motion, facial expressions, hands and arms motion. Section VI gives results of the recognition tests and Section VII concludes the report and outlines future directions.

## II. SIGN LANGUAGE TUTOR

### A. Sign Language

The linguistic characteristics of sign language is different than that of spoken languages due to the existence of several components affecting the context such as the use of facial expressions and the head movements in addition to the hand movements. The structure of spoken language makes use of words linearly i.e., one after another, whereas sign language makes use of several body movements in parallel in a completely different spatial and temporal sequence.

Language modeling enables to improve the performance of speech recognition systems. A language model for sign language is also required for the same purpose. Besides, the significance of co-articulation effects necessitates the continuous recognition of sign language instead of the recognition of isolated signs. These are complex problems to be tackled. For the present, we have focused on recognition of isolated words, or phrases, that involve manual and non-manual components.

There are many sign languages in the world. We have chosen signs from American Sign Language (ASL), since ASL is widely studied. However, our system is quite general and can be adapted to others.

### B. Database

For our database, 19 signs from American Sign Language were used. The selected signs include non-manual signs and inflections in the signing of the same manual sign [5]. For each sign, we recorded five repetitions from eight subjects. The preferred video resolution was 640*480 pixels and the frame rate was 25 fps. Short descriptions about the signs we used in the database can be seen in TABLE I.

TABLE I. ASL SIGNS IN THE DATABASE

| Sign | Head / Facial Expression | Hand |
|---|---|---|
| [smbdy] is here | Nod | Circular motion parallel to the ground with right hand. |
| Is [smbdy] here? | Brows up, Head forward | |
| [smbdy] is not here | Head shake | |
| Clean | - | Right palm facing down, left palm facing up. Sweep left hand with right. |
| Very clean | Lips closed, head turns from right to frontt, sharp motion | |
| Afraid | - | Hands start from the sides and meet in front of body, in the middle |
| Very afraid | Facial expression (lips open, eyes wide) | The same as "afraid", but shake the hands at the middle |
| Fast | - | Hands start in front of body and motion towards the body. Fingers partially closed, thumb open |
| Very fast | Facial expression (lips open, eyes wide), and sharp motion | |
| To drink | Head motion (up and down) | Drinking motion, hand as holding a cup |
| Drink (noun) | - | Repetitive drinking motion, hand as holding a cup. |
| To open door | - | Palms facing to the front. One hand moves as if the door is opened; only once. |
| Open door (noun) | - | Palms facing to the front. One hand moves as if the door is opened. Repeat, with small hand motion |
| Study | - | Left hand palm facing upwards, right hand all fingers open, mainly finger motion (finger tilt) |
| Study continuously | Circular head motion accompanies hand motion | Left palm facing up, right hand all fingers open, finger tilt together with large and downward circular motion |
| Study regularly | Downward head motion accompanies hand motion | Left palm facing upwards, right hand all fingers open, downward/ upward sharp motion, no finger motion |
| Look at | - | Starting from the eyes, forward motion, two hands together. |
| Look at continuously | Circular head motion accompanies hand motion | Starting from the eyes, forward motion, two hands together. Larger and circular motion |
| Look at regularly | Downward head motion accompanies hand motion | Starting from the eyes, forward motion, two hands together. Sharp forward/ backward motion |

### C. Tutor Application

The sign language tutor application was designed to show selected signs to the user and to let the user record his/her own sign using the webcam connected to the system. The graphical user interface for the tutor can be observed in Fig. 2.

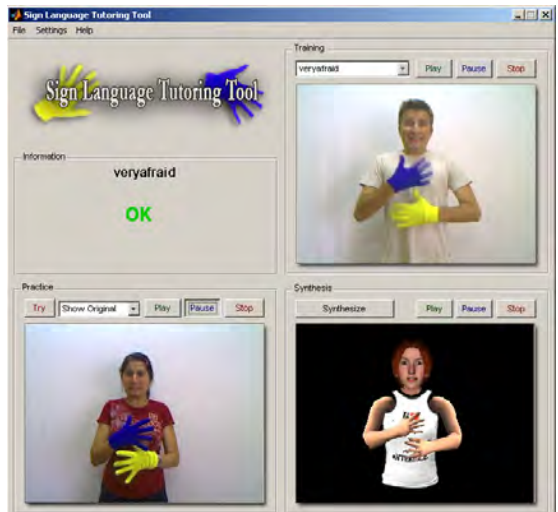

Fig. 2: Sign Language Tutoring Tool GUI

The graphical user interface consists of four panels: Training, Information, Practice and Synthesis. Training panel involves the teacher videos, thus the user can watch the videos to learn the sign by pressing the Play button. The program captures the user's sign video after the Try button is pressed. Afterwards, information panel is used for informing the user about the results of his/her trial. There are three types of results: "ok" (the sign was confirmed), "false" (the sign was wrong) and "head is ok but hands are false". Possible errors are also shown in this field.

Users can watch the original captured video or the segmented video in this panel as shown in Fig. 3. Afterwards, if the user wants to see the synthesized video, he/she can use the synthesis panel.

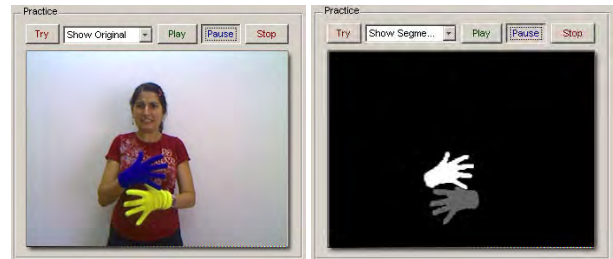

Fig. 3: A screenshot of original and segmented videos

### III. SIGN LANGUAGE ANALYSIS

#### A. Hand segmentation

The user wears gloves with different colors when performing the signs. The two colored regions are detected and marked as separate components. Ideally, we expect an image with three components: the background, the right hand and the left hand.

For the classification, histogram approach is used as proposed in [7]. Double thresholding is used to ensure connectivity, and to avoid spikes in the binary image. We prefer HSV color space as Jayaram et al. [7] and Albiol et al. [8] propose. HSV is preferred because of its robustness to changing illumination conditions.

The scheme is composed of training the histogram and threshold values for future use. We took 135 random snapshot images from our training video database. For each snapshot, ground truth binary images were constructed for the true position of the hands. Using the ground truth images, we have constructed the histogram for the left and right hands, resulting in two different histograms. Finally, normalization is needed for each histogram such that the values lie in the interval [0,1].

The low and high threshold values for double thresholding are found in training period. When single thresholding is used, a threshold value is chosen according to the miss and false alarm rates. Since we use double thresholding, we use an iterative scheme to minimize total error. We iteratively search for the minimum total error. This search is done in the range [μ-δ,μ+δ] to decrease the running time, where μ is the mean and δ is the standard deviation of the histogram.

After classification by using the scheme described above,

we observed that some confusing colors on the subject's clothing were classified as hand pixels. To avoid this, we selected the largest connected component of the classified regions into consideration. Thus we had only one component classified as hand for each color.

This classification approach can also be used for different colored gloves or skin after changing the ground truth images in the training period.

*B. Hand motion analysis*

The analysis of hand motion is done by tracking the center of mass (CoM) and calculating the velocity of each segmented hand. However, these hand trajectories are noisy due to noise introduced at the segmentation step. Thus, we use Kalman filters to smooth the obtained trajectories. The motion of each hand is approximated by a constant velocity motion model, in which the acceleration is neglected.

Two independent Kalman filters are used for each hand. The initialization of the Kalman Filter is done when the hand is first detected in the video. At each sequential frame, Kalman filter time update equations are calculated to predict the new hand position. The hand position found by the hand segmentation is used as measurements to correct the Kalman Filter parameters. Posterior states of each Kalman filter is defined as feature vectors for x, y coordinates of CoM and velocity. The hand can be lost due to occlusion or bad lighting in some frames. In that case, Kalman Filter prediction is directly used without correcting the Kalman Filter parameters. The hand is assumed to be out of the camera view if no hand can be detected for some number of (i.e. six) consecutive frames. Fig. 4 shows the extracted trajectories for each hand for the "fast" sign.

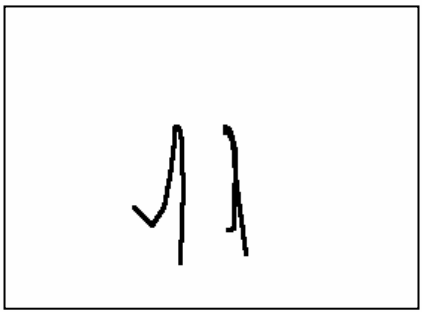

Fig. 4. Hand trajectories for sign "fast"

*C. Hand shape analysis*

Hand shape analysis is performed during sign recognition in order to increase the accuracy of recognition system and to differentiate between signs that differ only in hand shape. Each sign has a specific movement of the head, hands and hand postures. The extreme situation is when two signs have the same movements of head and hands and they differ only in hand postures. In this case, hand shape analysis is necessary to distinguish between them.

Another application can be in sign synthesis. If we analyze an unknown gesture and want to synthesize it with the same movements to caricature the movements of the actor, then finger and palm movements may be synthesized by following these steps: 1) unknown hand shape is classified into one of predefined clusters, 2) hand posture synthesis of classified cluster is performed (synthesis is prepared for each cluster). This can be useful whenever it is difficult to analyze finger and palm positions directly from image, for example when only low resolution images are available. This was the case in this project – each hand shape image was smaller than 80x80 pixels.

*1) Input – binary image*

After the segmentation of the source image is done, two binary images (only two colours representing background and hand) of left and right hand are analyzed.The mirror reflection of the right hand is taken so we analyze both hands in the same geometry; with thumb to the right. There are several difficulties using these images:
1. Low resolution (max. 80 pixels wide in our case)
2. Segmentation errors due to blurring caused by fast movement (see Fig. 5b)
3. Two different hand postures can have the same binary image (see Fig. 5a; which can be left hand observed from top or right hand from bottom)

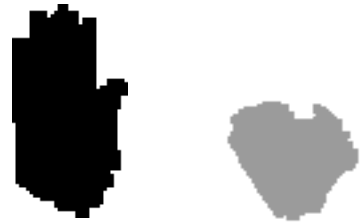

Fig.5. Two different hand segmentations: a. Hand shape 1; b. hand shape 2

*2) Hand shape analysis – feature extraction*

The binary image is converted into a set of numbers which describe hand shape, yielding the feature set. The aim is to have similar values of features for similar hand shapes and distant values for different shapes. It is also required to have scale invariant features so that images with the same hand shape but different size would have the same feature values. This is done by choosing features which are scale invariant. Our system uses only a single camera and our features do not have depth information; except for the foreshortening due to perspective. In order to keep this information about the z-coordinate (depth), five of the 19 features wer not normalized. All 19 features are listed in TABLE II.

TABLE II. HAND SHAPE FEATURES

| # | feature | invariant | |
|---|---|---|---|
| | | *scale* | *rotation* |
| 1 | Best fitting ellipse width | | ✓ |
| 2 | Best fitting ellipse height | | ✓ |

|    |                                                          | | *invariant* |
|----|----------------------------------------------------------|---|---|
| 3  | Compactness (perimeter²/area)                            | ✓ | ✓ |
| 4  | Ratio of hand pixels outside / inside of ellipse         | ✓ | ✓ |
| 5  | Ratio of hand / background pixels inside of ellipse      | ✓ | ✓ |
| 6  | sin (2*α)    α = angle of ellipse major axis             | ✓ |   |
| 7  | cos (2*α)    α = angle of ellipse major axis             | ✓ |   |
| 8  | Elongation (ratio of ellipse major/minor axis length)    | ✓ | ✓ |
| 9  | Percentage of NW (north-west) area filled by hand        | ✓ |   |
| 10 | Percentage of N area filled by hand                      | ✓ |   |
| 11 | Percentage of NE area filled by hand                     | ✓ |   |
| 12 | Percentage of E area filled by hand                      | ✓ |   |
| 13 | Percentage of SE area filled by hand                     | ✓ |   |
| 14 | Percentage of S area filled by hand                      | ✓ |   |
| 15 | Percentage of SW area filled by hand                     | ✓ |   |
| 16 | Percentage of W area filled by hand                      | ✓ |   |
| 17 | Total area (pixels)                                      |   | ✓ |
| 18 | Bounding box width                                       |   |   |
| 19 | Bounding box height                                      |   |   |

An initial idea was to use "high level" knowledge about the shape such as finger count, but the problems listed previously caused us to use more low level features, which are robust to segmentation errors and work well with low resolution images.

Seven of the features (#1,2,4,5,6,7,8) are based on using the best fitting ellipse (in least-squares sense) to a binary image, as seen in Fig. 6a.. The angle α is a value from 0° to 360°. However, only only values from 0 to 180 are meaningful, because the ellipse has mirror symmetry. Hence only 0° to 180° interval is used. Another problem is the following: Consider 5° and 15° ellipses, which have similar angles and similar orientation. 5° and 175° ellipses have similar orientations as before, but the angles are completely different. In order to represent this difference, we use sin(2*α) and cos(2*α) as features.

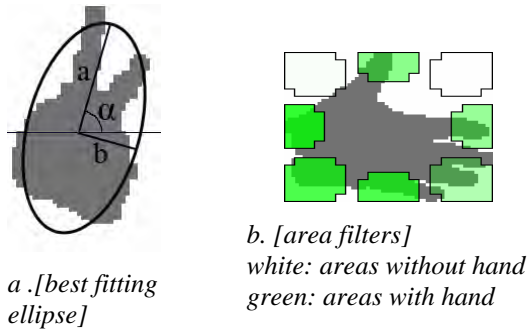

*a .[best fitting ellipse]*

*b. [area filters]*
*white: areas without hand*
*green: areas with hand*

Fig. 6  a. Best fitting ellipse; b. Area filters

Features #9 to 16 are based on using "area filters", as seen in Fig. 6a. The bounding box of the hand is divided into eight areas, in which percentage of hand pixels are calculated.

Other features in TABLE II, are perimeter, area and bounding box width and height.

*3) Classification*

Hand shape classification can be used for sign synthesis or to improve the recognition: The classified cluster can be used as new feature: We can use hand features for recognition only when the unknown hand shape is classified into a cluster (this means that the unknown hand shape is similar to a known one and not to a blurred shape which can have misleading features).

We have tried classification of hand shapes into 20 clusters (see Fig. 7 "clusters"). Each cluster is represented by approximately 15 templates. We use K-means algorithm (K=4) to classify unknown hand shape (represented by set of features described above). If the distance of unknown shape and each cluster is greater than 0.6 then this shape is declared as unclassified.

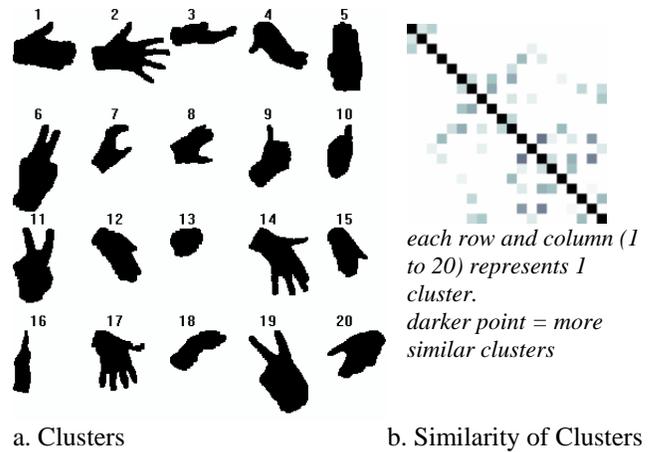

*each row and column (1 to 20) represents 1 cluster.*
*darker point = more similar clusters*

a. Clusters                    b. Similarity of Clusters

Fig. 7. a.The hand clusters; b.Similarity of clusters

As seen in Fig. 7b, some of the clusters are more similar than the others. For example, clusters 12, 14 and 19  are similar; so it is more difficult to correctly classify the unknown shape into one of these clusters.

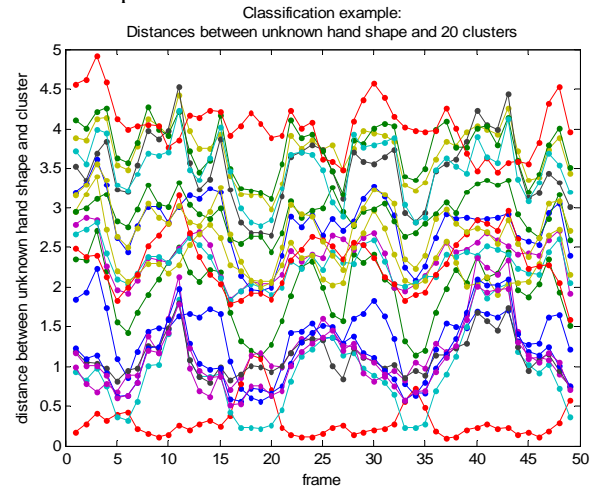

Fig. 8 Classification example: distances between an unknown hand shape and center of 20 clusters.

Classification of hand shapes is made in each frame of video sequence. It is reasonable to use information from previous frames, because hand shape cannot change so fast in each frame (1 frame = 40ms). Usually the classification is the same as in the previous frame, as seen in Fig. 8, where an unknown shape is classified into a cluster with the smallest distance.

To avoid fast variations of classifications we proposed a filter which smoothes these distances by weighted averaging Fig. 9 shows a classification example with filtering.

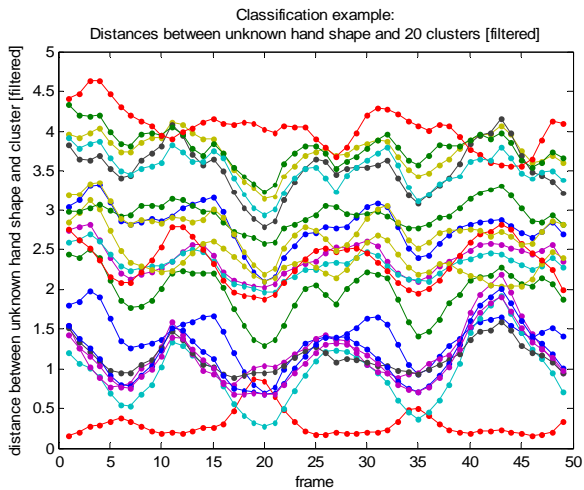

Fig. 9 Classification example: distances between an unknown hand shape and center of 20 clusters (filtered).

The new distance is calculated by the following equation:

$$D_{new}(t) = 0.34 \cdot D_{old}(t) + 0.25 \cdot D_{old}(t-1) + 0.18 \cdot D_{old}(t-2) + 0.12 \cdot D_{old}(t-3) + 0.07 \cdot D_{old}(t-4) + 0.04 \cdot D_{old}(t-5)$$

By comparing Fig. 8 and 9, one can see that this filter prevents fast changes in frames 5 and 6. This filter is designed to work in real-time applications. If used in offline application, it can easily be changed to use information from the future to increase the accuracy.

### D. Head motion analysis

*1) General Overview of the system*

Once a bounding box around the sign language student's face has been detected, rigid head motions such as head rotations and head nods are detected by using an algorithm working in a way close to the human visual system. In a first step, a filter inspired by the modeling of the human retina is applied. This filter enhances moving contours and cancels static ones. In a second step, the fast fourier transform (FFT) of the filtered image is computed in the log polar domain as a model of the primary visual cortex (V1). This step allows extracting two types of features: the quantity of motion and motion event alerts. In parallel, an optic flow algorithm extracts both vertical and velocity information only on the motion events alerts provided by the visual cortex stage. Fig. 10 gives a general overview of the algorithm. This module provides three features per frame: the quantity of motion, horizontal velocity and vertical velocity.

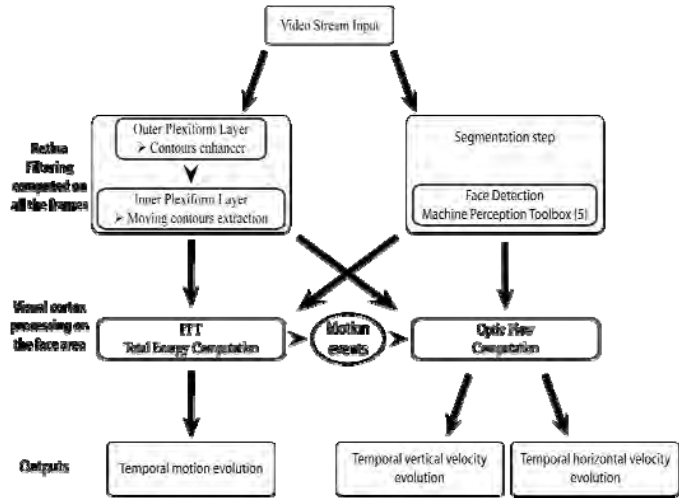

Fig. 10: Algorithm for rigid head motion data extraction

*2) Description of the components*

The first step consists in an efficient prefiltering [9]: the retina OPL (Outer Plexiform Layer) that enhances all contours by attenuating spatio-temporal noise, correcting luminance and whitening the spectrum (see Fig. 2). The IPL filter (Inner Plexiform Layer) [9] removes the static contours and extracts moving ones. This prefiltering is essential for data enhancement and allows minimizing the common problems of video acquisition such as luminance variations and noise.

The second step consists in a frequency analysis of the IPL filter output around the face whose response is presented on Fig.11. By computing the total energy of the amplitude spectrum of this output, as described in [10], we have information that depends linearly on the motion. The temporal evolution of this signal is the first data that is used in the sign language analyzer.

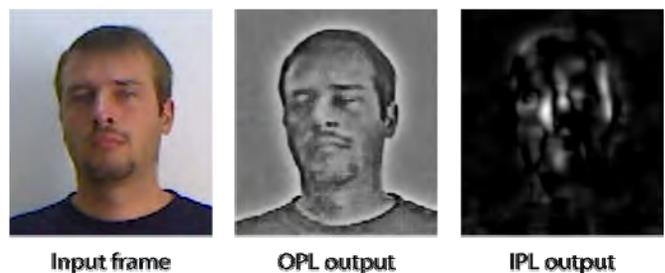

Fig. 11: Retina preprocessing outputs: extraction of enhanced contours (OPL) and moving contours (IPL)

In order to estimate the rigid head rotations [10], the proposed method analyses the spectrum of the IPL filter output in the log polar domain. It first detects head motion events [11] and is also able to extract its orientation. Then, in order to complete the description of the velocity, we propose to use features based on neuromorphic optical flow filters [12] which are oriented filters able to compute the velocity of the global head. Finally, optical flow is computed only when motion alerts are provided and its orientation is compared to the result given by the spectrum analysis. If the information is redundant, then we extract the velocity value at each frame, either horizontal or vertical in order to simplify the system.

*3) Extracted data sample*

In the end, the head analyzer is able to provide three signals per frame, information related to the quantity of motion and the vertical and horizontal velocity values. Fig. 12 shows two examples of the evolution of these signals, first in the case of a sequence in which the person expresses an affirmative « Here », second in the case of the expression of the sign « Very clean ». For the first sign, the head motion is a sequence of vertical head nods. Then, the quantity of motion indicator shows a periodic variation of its values with high amplitude for maximum velocity. The vertical velocity presents non zero values only during motion and also exhibits a periodic variation. On the contrary, the horizontal velocity indicator remains at zero. The « Very clean » sign consists of two opposite horizontal head motions. The quantity of motion indicator exhibits them. This time, the horizontal motion reports the velocity sign and amplitude variations while the vertical velocity indicator remains at zero. On this last sequence, we can see that some false alarms can be generated at the velocity output level: For example, at frame 68, a false horizontal motion is detected, but since the value of the quantity of motion is low, this velocity should not be taken into account. This is the advantage of using two detection signals: the cortex analysis model helps the velocity analyzer.

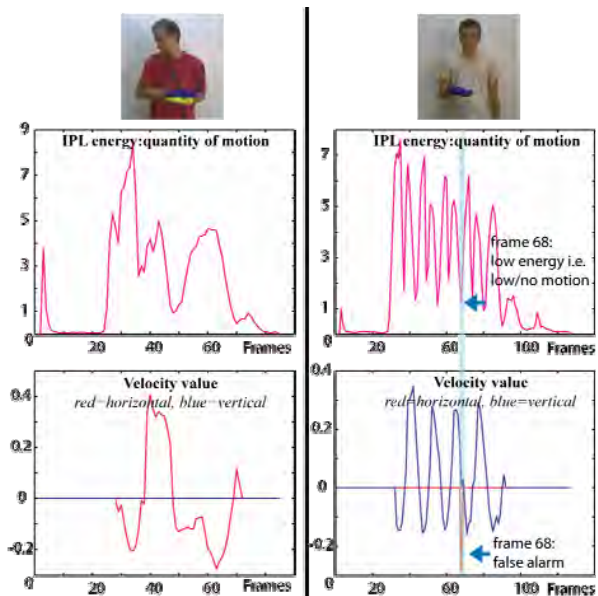

Fig. 12: Data extracted by the head analyzer

## IV. SIGN LANGUAGE RECOGNITION

### A. Preprocessing of sign sequences

The sequences obtained from the videos contain parts where the signer is not performing the sign (start and end parts) and some parts that can be considered as transition frames. These frames of the sequence are eliminated by looking at the result of the segmentation step:

- All the frames at the beginning of the sequence are eliminated until the hand is detected.
- If the hand can not be detected at the middle of the sequence for less than N frames, the shape information is copied from the last frame where there is detection.
- If the hand can not be detected for more than N consequent frames, the sign is assumed to be finished. Rests of the frames including the last N frames are eliminated.
- After indicating the start and end of the sequence and eliminating the unnecessary frames the transition frames can be eliminated by deleting T frames from the start and end of the sequence.

### B. Sign features and normalization issues

*1) Hand motion features*

The trajectories must be further normalized to obtain translation and scale invariance. We use a similar normalization strategy as in [13]. The normalized trajectory coordinates are calculated with the following formulas:

Let $(<x_1;y_1>;...;<x_t;y_t>;...;<x_N;y_N>)$ be the hand trajectory where N is the sequence length. For translation normalization, define $x_m$ and $y_m$:

$$x_m = (x_{max} + x_{min}) / 2$$
$$y_m = (y_{max} + y_{min}) / 2$$

where $x_m$ and $y_m$ are the mid-points of the range in x and y coordinates respectively. For scale normalization, define $d_x$ and $d_y$:

$$d_x = (x_{max} / x_{min}) / 2$$
$$d_y = (y_{max} / y_{min}) / 2$$

where $d_x$ and $d_y$ are the amount of spread in x and y coordinates respectively. The scaling factor is selected to be the maximum of the spread in x and y coordinates, since scaling with different factors disturbs the shape.

$$d = max(d_x; d_y)$$

The normalized trajectory coordinates, $(<x'_1;y'_1>;...;<x'_t;y'_t>;...;<x'_N;y'_N>)$ such that $0 <= x'_t, y'_t <= 1$, are then calculated as follows:

$$x'_t = 0.5 + 0.5 (x_t - x_m) / d$$
$$y'_t = 0.5 + 0.5 (y_t - y_m) / d$$

Since the signs can be also two handed, both hand trajectories must be normalized. However, normalizing the trajectory of the two hands independently may result in a possible loss of data. To solve this problem, the midpoints and the scaling factor of left and right hand trajectories are calculated jointly. Following this normalization step, the left and right hand trajectories are translated such that their starting position is (0,0).

*2) Hand position features*

In sign language, the position of the hand with respect to the body location is also important. We integrated position information by calculating the distance of the CoM of each hand to the face CoM. The distance at x and y coordinates are normalized by the face width and height respectively.

*3) Hand shape features*

All 19 hand shape features are normalized into values between 0 and 1. Features calculated as percentage (0 to 100%) are just divided by 100. The rest of features is normalized by using this equation:

$$F_{normalized} = (F - min) / (max - min)$$

where *min* is minimal value of feature (in training dataset) and *max* is maximum value. In case smaller or greater value occurs, $F_{normalized}$ is truncated to stay in $<0,1>$.

*4) Head motion features*

Head motion analysis provides three features that can be used in the recognition: motion energy of the head, horizontal and vertical velocity of the head. However these features are not invariant to differences that can exist between different performances of the same sign. Moreover, the head motion is not directly synchronized with the hand motion. To handle inter and intra personal differences, adaptive smoothing is applied to head motion features where α is used as 0.5:

$$F_i = \alpha F_i + (1 - \alpha) F_{i-1}$$

This smoothing has an effect of cancelling the noise between different performances of a sign and creating a smoother pattern.

*C. HMM modeling*

After sequence pre-processing and normalization, HMM models are trained for each sign, using Baum-Welch algorithm. We have trained 3 different HMMs for comparison purposes:

- $HMM_{manual}$ uses only hand information. Since hands form the basis of the signs, these models are expected to be very powerful in classification. However, absence of the head motion information prohibits a correct classification when the only difference of two signs is related to the head motion (i.e. here, ishere and not here)
- $HMM_{manual\&nonmanual}$ uses hand and head information. Since there is not a direct synchronization between hand and head motions, these models are not expected to have much better performance than $HMM_{manual}$. However using head information results in a slight increase in the performance.
- $HMM_{nonmanual}$ uses only head information. The head motion is complementary of the sign thus it can not be used alone to classify the signs. A data fusion methodology is needed to utilize these models together with models of manual components.

*D. Fusion of different modalities of sign language*

We have used a sequential score fusion strategy for combining manual and non-manual parts of the sign. We want our system to be as general as possible and capable of extending the sign set without changing the recognition system. Thus, we do not use any prior knowledge about the sign classes. For example, we know that here, *ishere* and *nothere* have exactly the same hand information but the head information differs. Using this prior information as a part of the recognition system increases the performance however the system looses its extendibility for upcoming signs since each sign will require a similar prior information. Instead we choose to extract the cluster information as a part of the recognition system.

Base decision is given by an HMM which uses both hand and head features in the same feature vector. However, the decision of these models is not totally correct since the head information is not utilized well. We used the likelihoods of $HMM_{nonmanual}$ to give the final decision.

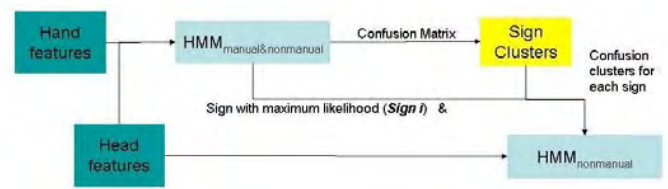

Fig. 13. Sequential fusion strategy

*1) Training*

- During training, models for each sign class are trained for $HMM_{manual\&nonmanual}$ and $HMM_{nonmanual}$.
- The cluster information for each sign is extracted from the confusion matrix of $HMM_{manual\&nonmanual}$. In the confusion matrix of the validation set the misclassifications are investigated. If all examples of a sign class are classified correctly, the cluster of that sign class only contains itself. For each misclassification, we add that sign class to the cluster.

*2) Testing*

The fusion strategy (Fig. 13) for an unseen test example is as follows:

- Likelihoods of $HMM_{manual\&nonmanual}$ for each sign class are calculated and the sign class with the maximum likelihood is selected as the base decision.
- Selected sign and its cluster information are sent to $HMM_{nonmanual}$.
- $HMM_{nonmanual}$ likelihood of the selected sign is calculated as well as the likelihoods of the signs in its cluster.
- Among these likelihoods, the sign class with the maximum $HMM_{nonmanual}$ likelihood id selected as the final decision.

## V. Synthesis and Animation

### A. Head motion and facial expression synthesis

The head synthesis performed in the present project is based on the MPEG-4 Facial Animation Standard [14], [15]. In order to ease the synthesis of a virtual face, the MPEG-4 Facial Animation (FA) defines two sets of parameters in a standardized way. The first set of parameters, the Facial Definition Parameter (FDP) set, is used to define 84 Feature Points (FP), located on morphological places of the neutral head, as depicted in Fig. 14 (black points). The feature points serve as anchors for 3D face deformable meshes, represented by a set of 3D vertices.

The second set defined by the MPEG-4 Standard is the Facial Animation Parameter (FAP) set. The Facial Animation Parameters (FAPs) represent a complete set of basic facial actions closely related to muscle movements and therefore allow the representation of facial expressions by modifying the positions of the previously defined feature points (FP). They consist of a set of 2 high-level (visemes and 6 archetypal emotions) and 66 low-level parameters (depicted as white filled points on Fig. 14). In this project, we only use the low-level parameters which are basic deformations applied to specific morphological places of the face, like the top middle outer-lip, the bottom right eyelid, etc...

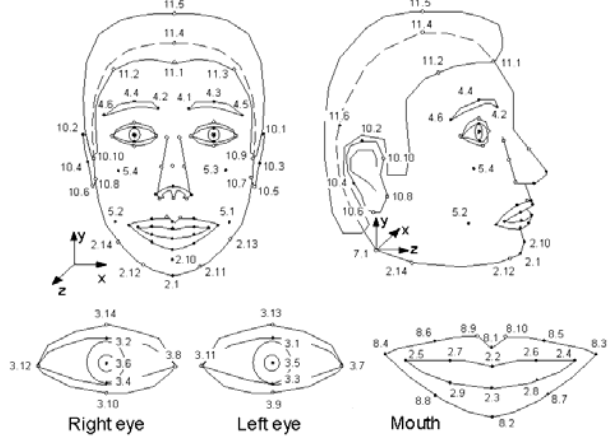

Fig. 14. *The 3D feature points of the FDP set*

The head synthesis system architecture is depicted on Fig. 15. As input, we receive the detected gesture (one data per sequence), the IPL energy, and the vertical and horizontal velocity of the head motion (as much data as frames in the sequence). We then filter and normalize these data in order to compute the head motion during the considered sequence. The result of the processing is expressed in terms of FAPs so that we can output a FAP file. The FAP file for the considered sequence is fed into the animation player. The animation player we used is an MPEG-4 compliant 3D talking head animation player developed by [16], part of an open source tools set available at [17]. Once rendered, we finally output an avi file containing the head synthesis sequence. Fig. 16 shows an example rendering with respect to the input data.

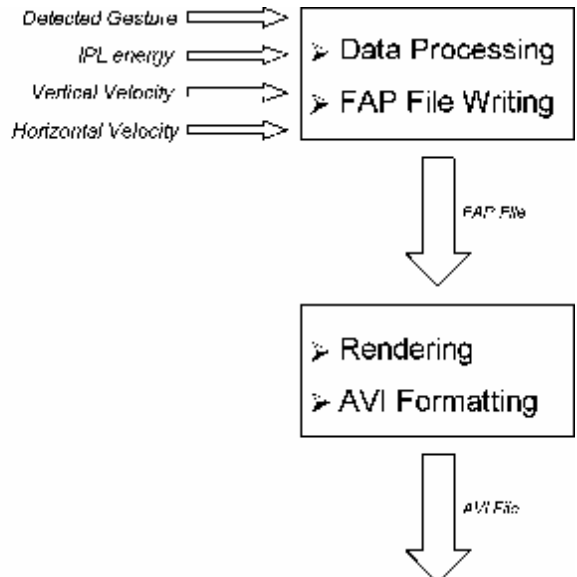

Fig. 15 : *Head synthesis system architecture*

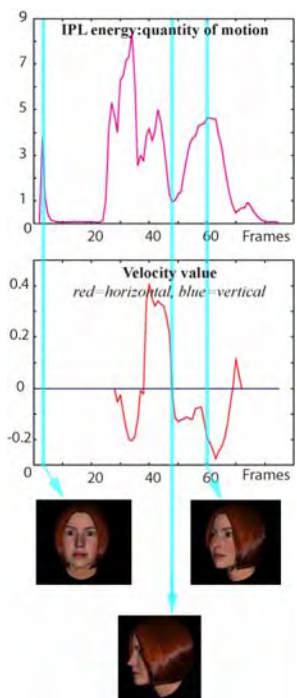

Fig. 16. Rendered head with different input values

### B. Hands and arms synthesis

For the hands and arms synthesis, we use a library of predefined positions for each gesture. Each animation has key positions (the positions that define the gesture) and interpolated positions. To create an animation we only need to set the key positions in the correct frames depending on the speed we want to get, and then interpolate the rest of the frames.

For our system, we need to do two different adaptations: speed adaptation, and position adaptation. The gesture detected is supposed to be closer to its predefined one, so we can define a set of steps that will be the same for each synthesis.

To create an animation from the features communicated by the analysis module, we have to follow the following steps:

*1. Physical features extraction:* from the input file, we extract the information about the head position, the length of the arms, and the maximal and minimal values of the hands coordinates (x,y). These will be used to normalize the information in order to adapt the system to our avatar features.

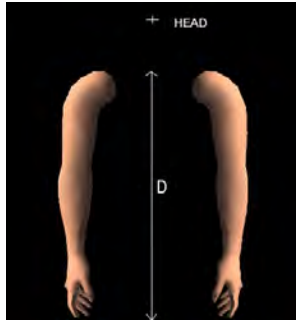

Fig. 17 : Arms system and main features.

*2. Speed features extraction:* For each gesture, we need to find the frames where the speed changes the most (border frames), because these frames will define our key positions. We find these features by differencing the coordinates of each frame and the next one. If the variation is smaller than a threshold we have set, it is supposed to be the same position than the last frame. It is necessary to know how much time we have to hold a position. With this extraction we have defined the frames where we will set the key positions.

*3. Physical adaptation and position definition:* We can adapt the the parameters of our avatar according to the analysis results. Depending on the minimal and maximal values, we have extracted for *x* and *y*, we choose two predefined positions for each border frame, and then we interpolate these positions to get a new one. All positions we get here will be our key positions.

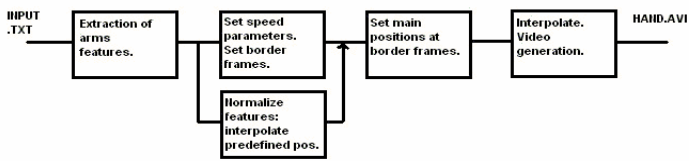

Fig. 18 : System structure for hands and arms synthesis.

*4. Set positions depending on the speed features:* To set the positions we have created, we take the border frames defined in step 2. We set the key positions for the border frames, holding them if necessary. After this, we only need to interpolate the rest of the frames, to get the final animation. From the final animation we will generate the video output that will be represented with the head result to show the complete avatar playing the gesture that the person in front of the camera did.

## VI. RECOGNITION RESULTS

We have used 70% of the signs in the database for training and the rest for testing. The distributions of sign classes are equal both in training and test sets. Confusion matrices and performance results are reported on the test set.

The confusion matrix of HMMs that are trained by using only hand information is shown in TABLE III. The total recognition rate is 67%. However, it can be seen that most of the misclassifications are between the sign groups where the hand information is the same or similar and the main difference is in the head information, which is not utilized in this scheme. When sign clusters are taken into account, there are only five misclassifications out of 228; resulting in a 97.8% recognition rate.

TABLE III. CONFUSION MATRIX. (ONLY HAND INFORMATION)

| Only Hand | door | to open | drink (noun) | to drink | here | is here? | not here | look at | look at cont. | look at reg. | study | study cont. | study reg. | afraid | very afraid | clean | very clean | fast | very fast |
|---|---|---|---|---|---|---|---|---|---|---|---|---|---|---|---|---|---|---|---|
| door | 10 | 2 | 0 | 0 | 0 | 0 | 0 | 0 | 0 | 0 | 0 | 0 | 0 | 0 | 0 | 0 | 0 | 0 | 0 |
| to open | 0 | 12 | 0 | 0 | 0 | 0 | 0 | 0 | 0 | 0 | 0 | 0 | 0 | 0 | 0 | 0 | 0 | 0 | 0 |
| drink (noun) | 0 | 0 | 12 | 0 | 0 | 0 | 0 | 0 | 0 | 0 | 0 | 0 | 0 | 0 | 0 | 0 | 0 | 0 | 0 |
| to drink | 0 | 0 | 1 | 11 | 0 | 0 | 0 | 0 | 0 | 0 | 0 | 0 | 0 | 0 | 0 | 0 | 0 | 0 | 0 |
| here | 0 | 0 | 0 | 0 | 4 | 3 | 5 | 0 | 0 | 0 | 0 | 0 | 0 | 0 | 0 | 0 | 0 | 0 | 0 |
| is here? | 0 | 0 | 0 | 0 | 0 | 5 | 7 | 0 | 0 | 0 | 0 | 0 | 0 | 0 | 0 | 0 | 0 | 0 | 0 |
| not here | 0 | 0 | 0 | 0 | 0 | 5 | 7 | 0 | 0 | 0 | 0 | 0 | 0 | 0 | 0 | 0 | 0 | 0 | 0 |
| look at | 0 | 0 | 0 | 0 | 0 | 0 | 0 | 7 | 1 | 4 | 0 | 0 | 0 | 0 | 0 | 0 | 0 | 0 | 0 |
| look at cont. | 0 | 0 | 0 | 0 | 0 | 0 | 0 | 0 | 12 | 0 | 0 | 0 | 0 | 0 | 0 | 0 | 0 | 0 | 0 |
| look at reg. | 3 | 1 | 0 | 0 | 0 | 0 | 0 | 0 | 1 | 7 | 0 | 0 | 0 | 0 | 0 | 0 | 0 | 0 | 0 |
| study | 0 | 0 | 0 | 0 | 0 | 0 | 0 | 0 | 0 | 0 | 4 | 4 | 4 | 0 | 0 | 0 | 0 | 0 | 0 |
| study cont. | 0 | 0 | 0 | 0 | 0 | 0 | 0 | 0 | 0 | 0 | 0 | 8 | 4 | 0 | 0 | 0 | 0 | 0 | 0 |
| study reg. | 0 | 0 | 0 | 0 | 0 | 0 | 0 | 0 | 0 | 0 | 1 | 1 | 10 | 0 | 0 | 0 | 0 | 0 | 0 |
| afraid | 0 | 0 | 0 | 0 | 0 | 0 | 0 | 0 | 0 | 0 | 0 | 0 | 0 | 2 | 10 | 0 | 0 | 0 | 0 |
| very afraid | 0 | 0 | 0 | 0 | 0 | 0 | 0 | 0 | 0 | 0 | 0 | 0 | 0 | 0 | 12 | 0 | 0 | 0 | 0 |
| clean | 0 | 0 | 0 | 0 | 0 | 0 | 0 | 0 | 0 | 0 | 0 | 0 | 0 | 0 | 0 | 6 | 6 | 0 | 0 |
| very clean | 0 | 0 | 0 | 0 | 0 | 0 | 0 | 0 | 0 | 0 | 0 | 0 | 0 | 0 | 0 | 2 | 10 | 0 | 0 |
| fast | 0 | 0 | 0 | 0 | 0 | 0 | 0 | 0 | 1 | 0 | 0 | 0 | 0 | 0 | 0 | 0 | 0 | 3 | 8 |
| very fast | 0 | 0 | 0 | 0 | 0 | 0 | 0 | 0 | 0 | 0 | 0 | 0 | 0 | 0 | 0 | 0 | 0 | 1 | 11 |

The confusion matrix of HMMs that are trained by the combined feature vector of hand and head information is shown in TABLE IV. The total performance is 77%. All misclassifications, except one, are between the sign groups. Although there is a slight increase in the performance, this fusion method does not utilize the head information effectively. Therefore, we have adopted the sequential fusion fusion strategy described in Section IVD.

The confusion matrix of the sequential fusion methodology is shown in TABLE IV. The total performance is 85.5%. The misclassifications between the sign groups are very few except for the study and look at sign groups. The reason of these misclassifications can be related to the deficiency of vision hardware or to the misleading feature values:

- The *study* sign: The confusion between *study regularly* and *study continuously* can stem from a deficiency of the 2D capture system. These two signs differ mainly in the third dimension, which we cannot capture. The confusion between *study* and *study regularly* can be a result of over-smoothing the trajectory.
- For the *look at* sign, the hands can be in front of the head for many of the frames. For those frames, the face detector may fail to detect the face and may provide wrong feature values which can mislead the recognizer.

Manual sign classification performance is 99.5%, which means only one sign is misclassified out of 228.

TABLE IV. CONFUSION MATRIX. FEATURE LEVEL FUSION

| Hand Head feature fusion | door | to open | drink (noun) | to drink | here | is here? | not here | look at | look at cont. | look at reg. | study | study cont. | study reg. | afraid | very afraid | clean | very clean | fast | very fast |
|---|---|---|---|---|---|---|---|---|---|---|---|---|---|---|---|---|---|---|---|
| door | 11 | 1 | 0 | 0 | 0 | 0 | 0 | 0 | 0 | 0 | 0 | 0 | 0 | 0 | 0 | 0 | 0 | 0 | 0 |
| to open | 1 | 11 | 0 | 0 | 0 | 0 | 0 | 0 | 0 | 0 | 0 | 0 | 0 | 0 | 0 | 0 | 0 | 0 | 0 |
| drink (noun) | 0 | 0 | 12 | 0 | 0 | 0 | 0 | 0 | 0 | 0 | 0 | 0 | 0 | 0 | 0 | 0 | 0 | 0 | 0 |
| to drink | 0 | 0 | 0 | 12 | 0 | 0 | 0 | 0 | 0 | 0 | 0 | 0 | 0 | 0 | 0 | 0 | 0 | 0 | 0 |
| here | 0 | 0 | 0 | 0 | 4 | 4 | 4 | 0 | 0 | 0 | 0 | 0 | 0 | 0 | 0 | 0 | 0 | 0 | 0 |
| is here? | 0 | 0 | 0 | 0 | 0 | 12 | 0 | 0 | 0 | 0 | 0 | 0 | 0 | 0 | 0 | 0 | 0 | 0 | 0 |
| not here | 0 | 0 | 0 | 0 | 0 | 2 | 10 | 0 | 0 | 0 | 0 | 0 | 0 | 0 | 0 | 0 | 0 | 0 | 0 |
| look at | 0 | 0 | 0 | 0 | 0 | 0 | 0 | 7 | 1 | 4 | 0 | 0 | 0 | 0 | 0 | 0 | 0 | 0 | 0 |
| look at cont. | 0 | 0 | 0 | 0 | 0 | 0 | 0 | 0 | 12 | 0 | 0 | 0 | 0 | 0 | 0 | 0 | 0 | 0 | 0 |
| look at reg. | 1 | 0 | 0 | 0 | 0 | 0 | 0 | 0 | 4 | 7 | 0 | 0 | 0 | 0 | 0 | 0 | 0 | 0 | 0 |
| study | 0 | 0 | 0 | 0 | 0 | 0 | 0 | 0 | 0 | 0 | 3 | 0 | 9 | 0 | 0 | 0 | 0 | 0 | 0 |
| study cont. | 0 | 0 | 0 | 0 | 0 | 0 | 0 | 0 | 0 | 0 | 0 | 8 | 4 | 0 | 0 | 0 | 0 | 0 | 0 |
| study reg. | 0 | 0 | 0 | 0 | 0 | 0 | 0 | 0 | 0 | 0 | 0 | 2 | 10 | 0 | 0 | 0 | 0 | 0 | 0 |
| afraid | 0 | 0 | 0 | 0 | 0 | 0 | 0 | 0 | 0 | 0 | 0 | 0 | 0 | 3 | 9 | 0 | 0 | 0 | 0 |
| very afraid | 0 | 0 | 0 | 0 | 0 | 0 | 0 | 0 | 0 | 0 | 0 | 0 | 0 | 0 | 12 | 0 | 0 | 0 | 0 |
| clean | 0 | 0 | 0 | 0 | 0 | 0 | 0 | 0 | 0 | 0 | 0 | 0 | 0 | 0 | 0 | 11 | 1 | 0 | 0 |
| very clean | 0 | 0 | 0 | 0 | 0 | 0 | 0 | 0 | 0 | 0 | 0 | 0 | 0 | 0 | 0 | 2 | 10 | 0 | 0 |
| fast | 0 | 0 | 0 | 0 | 0 | 0 | 0 | 0 | 0 | 0 | 0 | 0 | 0 | 0 | 0 | 0 | 0 | 6 | 6 |
| very fast | 0 | 0 | 0 | 0 | 0 | 0 | 0 | 0 | 0 | 0 | 0 | 0 | 0 | 0 | 0 | 0 | 0 | 0 | 12 |

TABLE 3. CONFUSION MATRIX. SEQUENTIAL FUSION

| Hand Head sequential fusion | door | to open | drink (noun) | to drink | here | is here? | not here | look at | look at cont. | look at reg. | study | study cont. | study reg. | afraid | very afraid | clean | very clean | fast | very fast |
|---|---|---|---|---|---|---|---|---|---|---|---|---|---|---|---|---|---|---|---|
| door | 11 | 1 | 0 | 0 | 0 | 0 | 0 | 0 | 0 | 0 | 0 | 0 | 0 | 0 | 0 | 0 | 0 | 0 | 0 |
| to open | 1 | 11 | 0 | 0 | 0 | 0 | 0 | 0 | 0 | 0 | 0 | 0 | 0 | 0 | 0 | 0 | 0 | 0 | 0 |
| drink (noun) | 0 | 0 | 12 | 0 | 0 | 0 | 0 | 0 | 0 | 0 | 0 | 0 | 0 | 0 | 0 | 0 | 0 | 0 | 0 |
| to drink | 0 | 0 | 0 | 12 | 0 | 0 | 0 | 0 | 0 | 0 | 0 | 0 | 0 | 0 | 0 | 0 | 0 | 0 | 0 |
| here | 0 | 0 | 0 | 0 | 11 | 0 | 1 | 0 | 0 | 0 | 0 | 0 | 0 | 0 | 0 | 0 | 0 | 0 | 0 |
| is here? | 0 | 0 | 0 | 0 | 0 | 11 | 1 | 0 | 0 | 0 | 0 | 0 | 0 | 0 | 0 | 0 | 0 | 0 | 0 |
| not here | 0 | 0 | 0 | 0 | 0 | 0 | 12 | 0 | 0 | 0 | 0 | 0 | 0 | 0 | 0 | 0 | 0 | 0 | 0 |
| look at | 0 | 0 | 0 | 0 | 0 | 0 | 0 | 10 | 0 | 2 | 0 | 0 | 0 | 0 | 0 | 0 | 0 | 0 | 0 |
| look at cont. | 0 | 0 | 0 | 0 | 0 | 0 | 0 | 2 | 9 | 1 | 0 | 0 | 0 | 0 | 0 | 0 | 0 | 0 | 0 |
| look at reg. | 1 | 0 | 0 | 0 | 0 | 0 | 0 | 5 | 4 | 2 | 0 | 0 | 0 | 0 | 0 | 0 | 0 | 0 | 0 |
| study | 0 | 0 | 0 | 0 | 0 | 0 | 0 | 0 | 0 | 0 | 9 | 3 | 0 | 0 | 0 | 0 | 0 | 0 | 0 |
| study cont. | 0 | 0 | 0 | 0 | 0 | 0 | 0 | 0 | 0 | 0 | 9 | 3 | 0 | 0 | 0 | 0 | 0 | 0 | 0 |
| study reg. | 0 | 0 | 0 | 0 | 0 | 0 | 0 | 0 | 0 | 0 | 0 | 5 | 7 | 0 | 0 | 0 | 0 | 0 | 0 |
| afraid | 0 | 0 | 0 | 0 | 0 | 0 | 0 | 0 | 0 | 0 | 0 | 0 | 0 | 12 | 0 | 0 | 0 | 0 | 0 |
| very afraid | 0 | 0 | 0 | 0 | 0 | 0 | 0 | 0 | 0 | 0 | 0 | 0 | 0 | 1 | 11 | 0 | 0 | 0 | 0 |
| clean | 0 | 0 | 0 | 0 | 0 | 0 | 0 | 0 | 0 | 0 | 0 | 0 | 0 | 0 | 0 | 12 | 0 | 0 | 0 |
| very clean | 0 | 0 | 0 | 0 | 0 | 0 | 0 | 0 | 0 | 0 | 0 | 0 | 0 | 0 | 0 | 0 | 12 | 0 | 0 |
| fast | 0 | 0 | 0 | 0 | 0 | 0 | 0 | 0 | 0 | 0 | 0 | 0 | 0 | 0 | 0 | 0 | 0 | 12 | 0 |
| very fast | 0 | 0 | 0 | 0 | 0 | 0 | 0 | 0 | 0 | 0 | 0 | 0 | 0 | 0 | 0 | 0 | 0 | 2 | 10 |

## VII. CONCLUSIONS AND FUTURE WORK

In this project, we have developed a sign tutor application that lets users learn and practice signs from a predefined library. The tutor application records the practiced signs; analyses the hand shapes and movements as well as the head movements, classifies the sign, and gives feedback to the user. The feedback consists of both text information and synthesized video, which shows the user a caricaturized version of his movements when the sign is correctly classified. Our performance tests yield a 99% recognition rate on signs involving manual gestures and 85% recognition rate on signs that involve both manual and non manual components, such as head movement and facial expressions.

## ACKNOWLEDGMENT

We thank Jakov Pavlek and Vjekoslav Levacic, who have volunteered to be in the sign database.

## APPENDIX : SOFTWARE NOTES

Since the individual parts in this project were coded in C, C++ and MATLAB, we preferred MATLAB to combine them for the tutor. MATLAB GUI was used to prepare the user interface.

We used the "Machine Perception Toolbox" [18] for head analysis. For HMM training, HMM routines in [19] are used. We also used "Intel Open Source Computer Vision Library" [20] routines in our project.